%% file: main.tex
\documentclass[10pt,twocolumn,letterpaper]{article}

\usepackage[pagenumbers]{cvpr} 
\usepackage[misc]{ifsym}

\input{preamble}

\input{tables/tables}

\definecolor{cvprblue}{rgb}{0.21,0.49,0.74}
\usepackage[pagebackref,breaklinks,colorlinks,citecolor=cvprblue]{hyperref}

\def\paperID{14888} 
\def\confName{CVPR}
\def\confYear{2024}

\title{Bootstrapping SparseFormers from Vision Foundation Models}

\author{
Ziteng Gao \quad \quad Zhan Tong \quad \quad Kevin Qinghong Lin \quad \quad Joya Chen \quad \quad Mike Zheng Shou\textsuperscript{\Letter} \\
Show Lab, National University of Singapore
}

\begin{document}

\maketitle

\begin{abstract}
    The recently proposed SparseFormer architecture provides an alternative approach to visual understanding by utilizing a significantly lower number of visual tokens via adjusting RoIs, greatly reducing computational costs while still achieving promising performance. However, training SparseFormers from scratch is still expensive, and scaling up the number of parameters can be challenging. In this paper, we propose to bootstrap SparseFormers from ViT-based vision foundation models in a simple and efficient way. Since the majority of SparseFormer blocks are the standard transformer ones, we can inherit weights from large-scale pre-trained vision transformers and freeze them as much as possible. Therefore, we only need to train the SparseFormer-specific lightweight focusing transformer to adjust token RoIs and fine-tune a few early pre-trained blocks to align the final token representation. In such a way, we can bootstrap SparseFormer architectures from various large-scale pre-trained models (e.g., IN-21K pre-trained AugRegs or CLIPs) using a rather smaller amount of training samples (e.g., IN-1K) and without labels or captions within just a few hours. As a result, the bootstrapped unimodal SparseFormer (from AugReg-ViT-L/16-384) can reach $84.9\%$ accuracy on IN-1K with only $49$ tokens, and the multimodal SparseFormer from CLIPs also demonstrates notable zero-shot performance with highly reduced computational cost without seeing any caption during the bootstrapping procedure. In addition, CLIP-bootstrapped SparseFormers, which align the output space with language without seeing a word, can serve as efficient vision encoders in multimodal large language models. 
    Code and models are available at \href{https://github.com/showlab/sparseformer}{https://github.com/showlab/sparseformer}
\end{abstract}
\blfootnote{ \Letter: Corresponding Author.}

\input{section/intro}

\input{section/related}

\input{section/method}

\input{section/exps}
\input{section/conclusion}

\appendix
\section{Appendix}
\section*{A.1. More Ablations on Bootstrapping Settings}
In Section 3, We have proposed to {\em truncate the leading, tune the middle, and freeze the ending pre-trained transformer blocks} to further reduce the compute and preserve the output embedding space of the foundation transformer to bootstrap from.
Here, we investigate the effect of this bootstrapping paradigm in Table~\ref{tab:ablationsettings}.

\begin{table}[h!]
    \centering
    \footnotesize
    \renewcommand\arraystretch{1.2}
    \setlength{\tabcolsep}{1.3pt}
    \begin{tabular}{l|ccccccc}
    model 
          & \makecell[c]{\#truncate\\ blocks}
          & \makecell[c]{\#tunable\\ blocks}
          & \makecell[c]{IN-1K \\ top-1 acc.}
          & FLOPs
          & \#Params
          \\\shline
    \rowcolor{Gray}\sfb/, default & $4$ & $4$ of $8$ & 82.5 & 3.8G & 86M \\
    \sfb/, all frozen & $4$ & $0$ of $8$  & 81.8 & 3.8G & 86M \\
    \sfb/, all tunable & $4$ & $8$ of $8$  & 82.4 & 3.8G & 86M \\
    \sfb/, w/o truncation & $0$ & $4$ of $12$  & 82.7 & 5.2G & 92M \\
    \hline
    \rowcolor{Gray}\sfl/, default & $8$ & $8$ of $16$  & 84.5 & 11.4G & 213M \\
    \sfl/, all tunable & $8$ & $16$ of $16$  & 84.4 & 11.4G & 213M \\
    \sfl/, w/o truncation & $0$ & $8$ of $24$ & 84.3 & 16.4G & 314M \\
    \end{tabular}
    \caption{Ablation on truncating, tuning, and freezing settings.}\label{tab:ablationsettings}
  \end{table}
As shown in the table, bootstrapping SparseFormers without tuning pre-trained transformer blocks (``all frozen'') leads to inferior results compared to ones that do tune.
This is expected since frozen pre-trained transformer blocks can not adapt to the output of the focusing transformer during the bootstrapping procedure.
However, going to the opposite extreme of making all pre-trained blocks tunable (``all tunable'') can also be lagging.
This may be because the frozen classifier relies on the structure of the well-preserved output embedding space in our bootstrapping setting. We believe that this is also true for vision language models.
Besides that, we observe that bootstrapping without truncating leading blocks can be very unstable, and lead to different effects on \sfb/ and \sfl/ but with much more FLOPs and parameters.
Therefore, we choose our truncating the leading, tuning the middle, and freezing the ending paradigm as our bootstrapping design due to the reduced computation and minimal tunable parameters.

\section*{A.2. Experiment Settings in Details}
We here describe more experiment details in the bootstrapping procedure.
The learning rate for tuning pre-trained transformer blocks is set to $0.1\times$ that of the focusing transformer to make the training more stable after the warm-up.
The focusing transformer in our designed SparseFormer variant performs the feature sampling first, then self attention between tokens, the feed-forward network, and then the RoI adjustment for each iteration, in contrast to the original SparseFormer which the self attention is performed first and the feature sampling then.
We use this reversed order to prioritize the self-attention interaction between different tokens with sampled features.
We use two-layered MLP to produce RoI adjusting deltas in the focusing transformer.

Different from the original SparseFormers that do not inject positional information into latent tokens, we inject RoI-based position encoding into tokens after every feature sampling operation in the focusing transformer to align with typical vision transformers.
Our adopted positional encoding is also sinusoidal but in a continuous form:
\begin{align*}
    \textsc{PE}_{v} = \left[\sin(\pi f_0v),\cos(\pi f_0v), \sin(\pi f_1v),...\right]\in\mathbb{R}^{d/4},
\end{align*}
where $f_i$ is the frequency term that evenly lies in the exponential space from $1$ to $f_\textrm{max}=128$ (there are $d/8$ frequency terms), $v\in[v_\textrm{left}, v_\textrm{top}, v_\textrm{right}, v_\textrm{bottom}]$ where each component is the normalized coordinate of a token RoI that lies in $[0, 1]$. The final positional encoding is  these four positional encoding parts concatenated:
\begin{align*}
    \textsc{PE} = \left[\textsc{PE}_\textrm{left}|\textsc{PE}_\textrm{top}|\textsc{PE}_\textrm{right}|\textsc{PE}_\textrm{bottom}\right]\in\mathbb{R}^d.
\end{align*} 

\section*{A.2. More Visualizations}
\begin{figure}[b]
    \includegraphics[width=\linewidth]{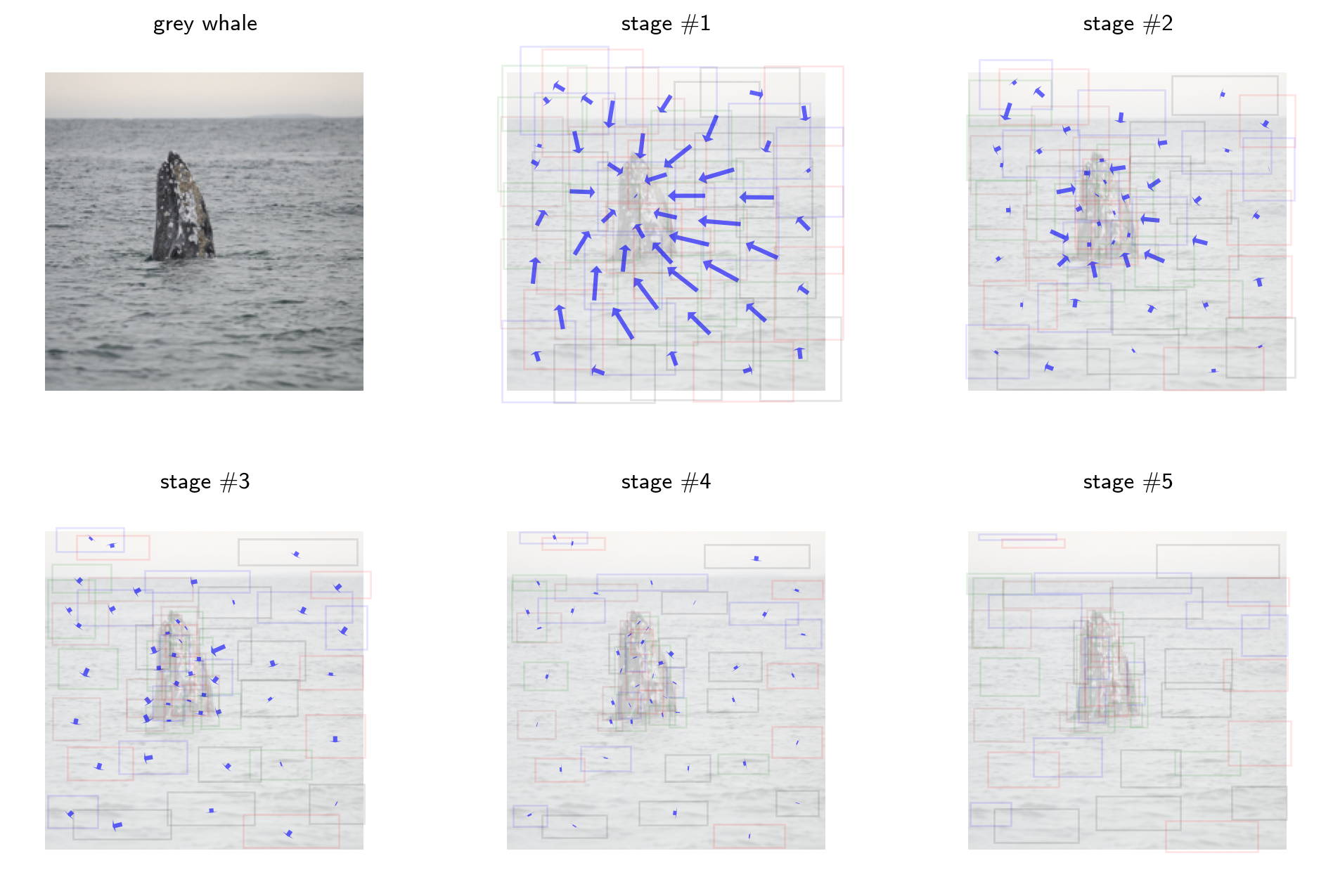}
    \caption{RoI adjustments in each iteration in \sfb/.}\label{adj1}
\end{figure}

\begin{figure*}[t]
    \centering
    \includegraphics[width=0.49\linewidth]{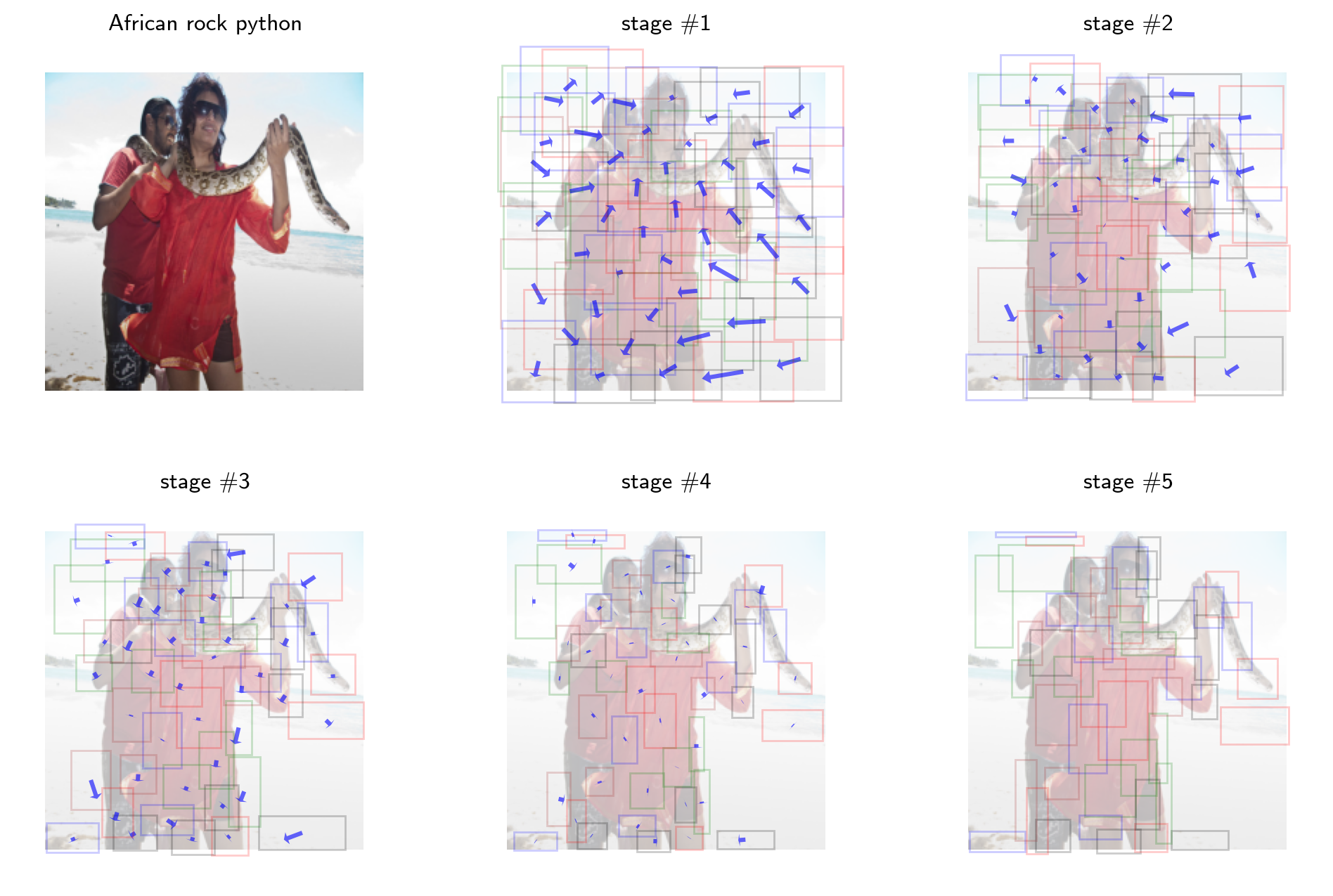} %
    \includegraphics[width=0.49\linewidth]{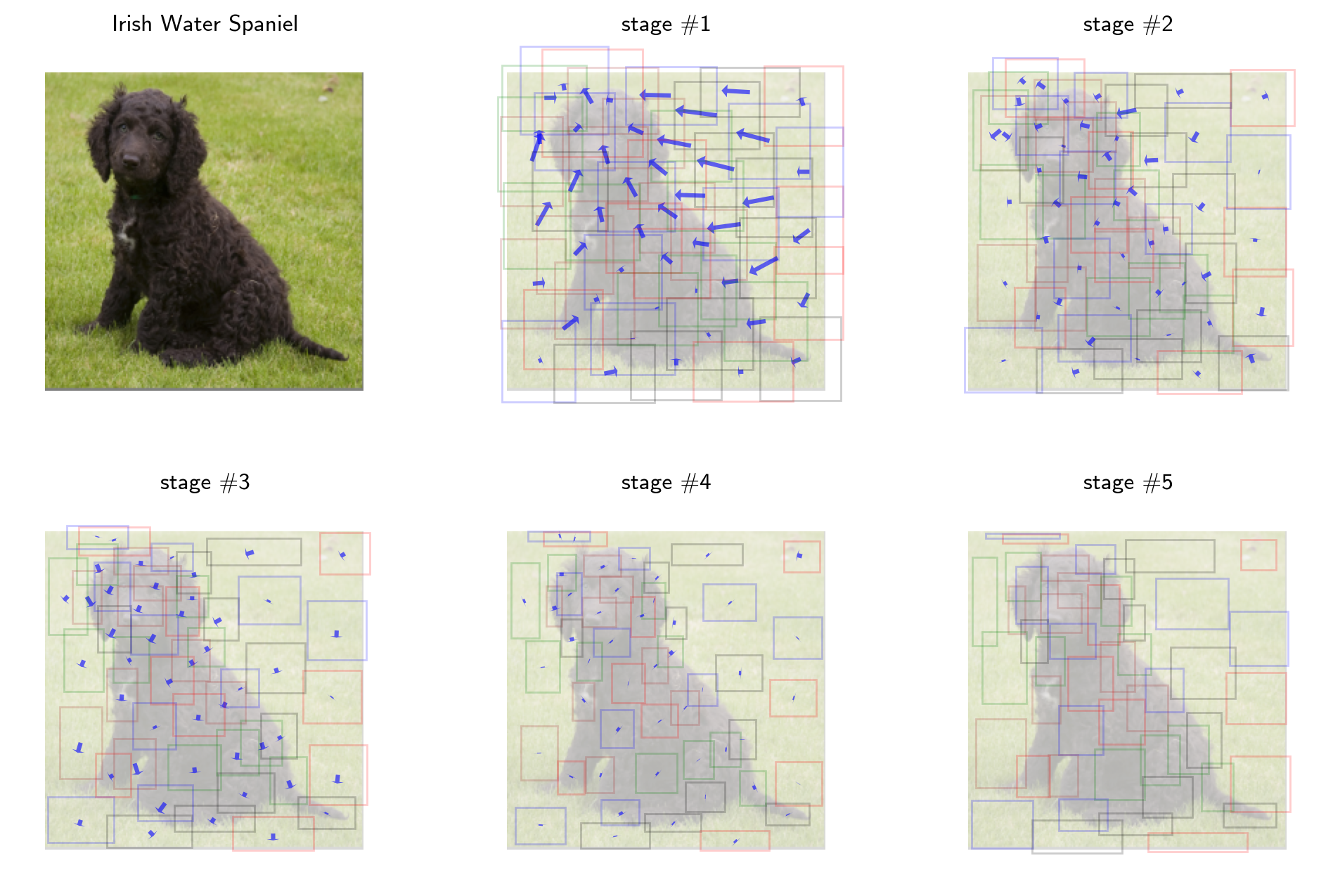}
    \caption{RoI adjustments (cont'd).}\label{adj2}
\end{figure*}

Here, we visualize the detailed RoI adjustments in each iteration in the focusing transformer of our bootstrapped SparseFormer \sfb/ in Figure~\ref{adj1} and \ref{adj2}.

\begin{figure*}[b]
    \centering
    \includegraphics[width=\textwidth]{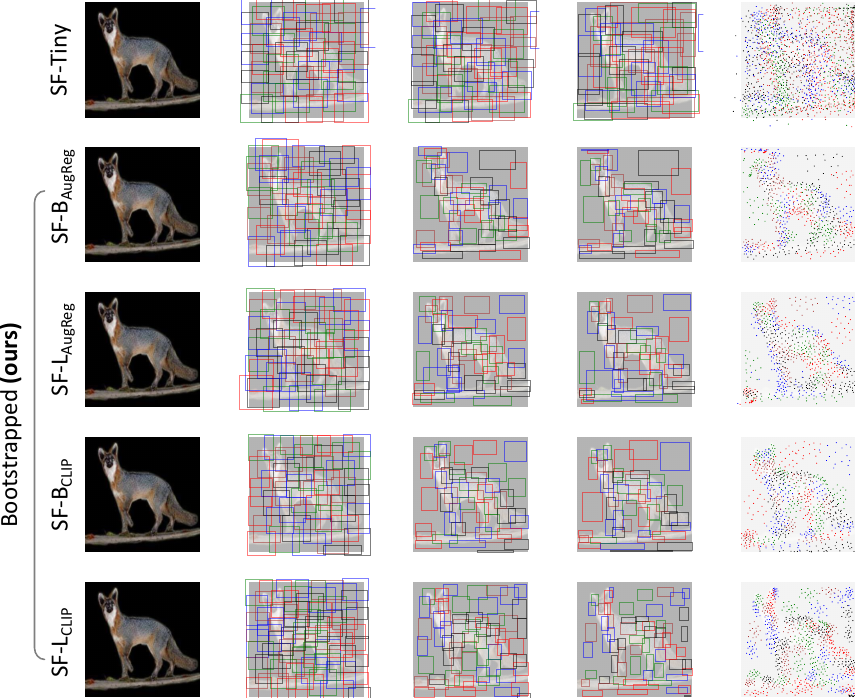}
    \caption{Visualizations on the original SparseFormer and our bootstrapped SparseFormers.
    For each image, there are an input image, token RoIs in the \{first, third, last\} stage, and sampling points in the last stage in the focusing transformer from left to right.}\label{com1}
\end{figure*}

\begin{figure*}[t]
    \centering
    \includegraphics[width=\textwidth]{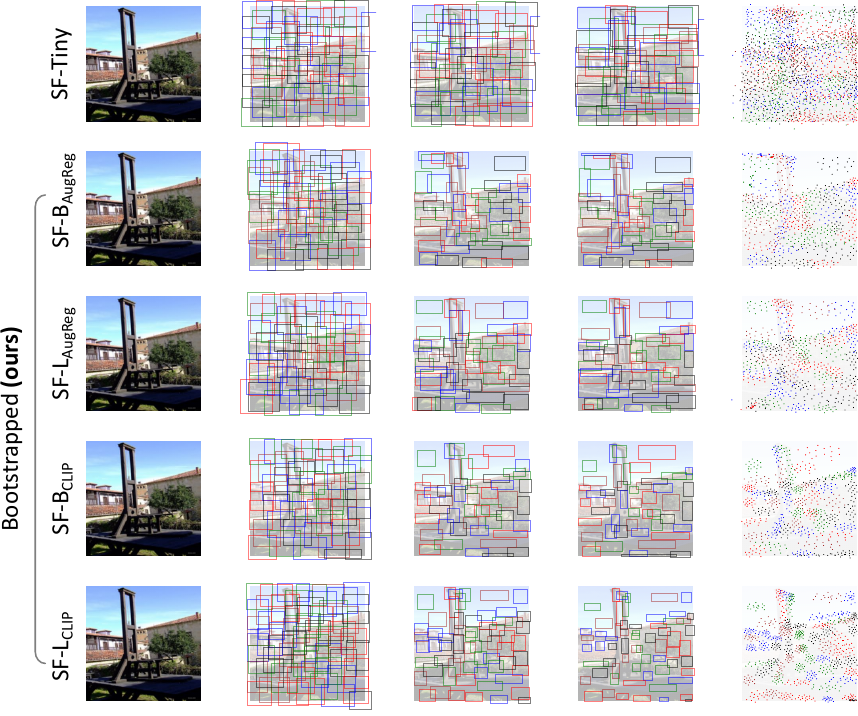}
    \caption{Visualizations (cont'd).}\label{com2}
\end{figure*}

In addition to that, we perform comprehensive visualizations on assorted bootstrapped SparseFormers (\sfb/, \sfl/, SF-B$_{\textrm{CLIP}}$, and SF-L$_{\textrm{CLIP}}$) on ImageNet-1K validation image samples in Figure~\ref{com1} and \ref{com2}.
Bootstrapped SparseFormers exhibit better sparsity and localization on foregrounds than the original SparseFormer.

\clearpage
\bibliography{iclr2024_conference}
\bibliographystyle{iclr2024_conference}
\end{document}

%% file: preamble.tex
\usepackage[dvipsnames]{xcolor}


%% file: tables/tables.tex
\usepackage{caption}
\usepackage{subcaption}
\usepackage{pifont} 
\usepackage{multirow, overpic, textpos, array}
\usepackage{makecell}

\usepackage{color, colortbl}
\definecolor{Gray}{gray}{0.92}
\definecolor{lightgray}{gray}{0.5}
\definecolor{light}{gray}{0.96}
\definecolor{lightgreen}{rgb}{0.0, 0.6, 0.2}
\usepackage{nicematrix}

\newlength\savewidth\newcommand\shline{\noalign{\global\savewidth\arrayrulewidth
  \global\arrayrulewidth 1pt}\hline\noalign{\global\arrayrulewidth\savewidth}}

\newcommand\blfootnote[1]{\begingroup\renewcommand\thefootnote{}\footnote{#1}\addtocounter{footnote}{-1}\endgroup}

\newcommand{\TableScienceQA}{
\begin{table}[b!]
    \centering
    \footnotesize
    \renewcommand\arraystretch{1.0}
    
    \begin{tabular}{l|cc}
        vision encoder & acc & multi-modal only acc \\\shline
        \rowcolor{Gray} ViT-L/16-224 & 91.2 & 88.6 \\
        SF-L$_{\textrm{CLIP}}$ w/ 64 tokens & 89.2 & 84.5 \\
        SF-L$_{\textrm{CLIP}}$ w/ 128 tokens & 89.8 & 85.8 \\
    \end{tabular}
    \caption{
    In-place vision encoder replacement in LLaVa on the ScienceQA test set.
    }\label{tab:scienceqa}
\end{table}
}

\def\augregb/{ViT-B/16-224$_\textrm{AugReg}$}
\def\maeb/{ViT-B/16-224$_\textrm{MAE}$}
\def\augregl/{ViT-L/16-224$_\textrm{AugReg}$}
\def\mael/{ViT-L/16-224$_\textrm{MAE}$}
\def\augreglup/{ViT-L/16-384$_\textrm{AugReg}$}
\def\sfb/{SF-B$_\textrm{AugReg}$}
\def\sfl/{SF-L$_\textrm{AugReg}$}
\newcommand{\TableConfiguration}{
\begin{table}[h]
  \centering
  \scriptsize
  \renewcommand\arraystretch{1.1}
  \setlength{\tabcolsep}{1.2pt}
  \begin{tabular}{l|ccccccc}
  model 
        & \#tokens
        & \makecell[c]{focusing\\ iterations}
        & \makecell[c]{focusing\\ dimension}
        & \makecell[c]{\#truncate\\ blocks}
        & \makecell[c]{\#tunable\\ blocks}
        & FLOPs
        & \#Params
        \\\shline
  \rowcolor{Gray} \augregb/ & 196 & - & - & - & - & 17.5G & 86M \\
  \sfb/ & 49 & 4+1 & 384 & $4$ & $4$ & 3.8G & 64M \\
  \rowcolor{Gray} \augregl/ & 196 & - & - & - & - & 61.6G & 304M \\
  \sfl/ & 49 & 4+1 & 512 & $8$ & $8$ & 11.4G & 213M \\
  \end{tabular}
  \caption{Configurations of SparseFormers.}\label{tab:configuration}
\end{table}
}

\newcommand{\TableUnimodalExp}{
\begin{table}[t]
  \centering
  \footnotesize
  \renewcommand\arraystretch{1.1}
  \setlength{\tabcolsep}{0.6pt}
  \begin{tabular}{l|ccll}
  model 
        & top-1 acc.
        & \#Params
        & FLOPs
        & img/s
        \\\shline
  \rowcolor{Gray} \augregb/, 196 tokens &
        84.6 & 86M & 17.5G & 1126 \\
  {\sfb/}, 49 tokens &
        82.5 & 64M & 3.8G$_{\color{lightgreen} \downarrow 78.3\%}$  & 3001$_{\color{lightgreen}\uparrow 166\%}$ \\
  \sfb/, 80 tokens$\uparrow$ &
        83.4 & 64M & 6.2G$_{\color{lightgreen}\downarrow 64.6\%}$  & 2080$_{\color{lightgreen}\uparrow 85\%}$ \\
  \sfb/, 80 tokens$\uparrow$, 320px$\uparrow$ &
        83.7 & 64M & 6.3G$_{\color{lightgreen}\downarrow 64.0\%}$  & 1898$_{\color{lightgreen}\uparrow 67\%}$ \\
  \hline
  \rowcolor{Gray}\augregl/, 196 tokens &
        85.8 & 304M & 61.6G & 388 \\
  {\sfl/}, 49 tokens &
        84.5 & 213M & 11.4G$_{\color{lightgreen}\downarrow 81.5\%}$  & 1557$_{\color{lightgreen}\uparrow 301\%}$ \\
  \sfl/, 80 tokens$\uparrow$ &
        85.2 & 213M & 18.6G$_{\color{lightgreen}\downarrow 69.8\%}$  & 1008$_{\color{lightgreen}\uparrow 160\%}$ \\
  \sfl/, 80 tokens$\uparrow$, 320px$\uparrow$ &
        85.5  & 213M & 18.7G$_{\color{lightgreen}\downarrow 69.6\%}$  & 928$_{\color{lightgreen}\uparrow 139\%}$ \\
  \hline
  \rowcolor{Gray}\augreglup/, 576 tokens &
        86.9 & 304M & 191G & 105 \\
  {\sfl/}, 49 tokens &
        84.9 & 213M & 11.6G$_{\color{lightgreen}\downarrow 93.9\%}$ & 1353$_{\color{lightgreen}\uparrow 1189\%}$ \\
  \sfl/, 144 tokens$\uparrow$ &
        86.5 & 213M & 33.9G$_{\color{lightgreen}\downarrow 82.3\%}$ & 560$_{\color{lightgreen}\uparrow 433\%}$ \\
  \sfl/, 196 tokens$\uparrow$ &
        86.7 & 213M & 46.5G$_{\color{lightgreen}\downarrow 75.7\%}$ & 378$_{\color{lightgreen}\uparrow 260\%}$ \\
  \end{tabular}
  \caption{
      Results of SparseFormers bootstrapped from AugReg models.
      The \textit{default} {SparseFormer} token number is~49.
      Throughputs are measured with half-precision data type and batch size of 128 on one A5000 GPU.
  }\label{tab:unimodal}
\end{table}
}

\newcommand{\TableComparisonWithTokenDropping}{
\begin{table}[t]
  \centering
  \footnotesize
  \renewcommand\arraystretch{1.1}
  \setlength{\tabcolsep}{2.6pt}
  \begin{tabular}{l|cll}
  model 
        & top-1 acc.
        & FLOPs
        & img/s
        \\\shline
  \rowcolor{Gray} \augregb/&
        84.6 & 17.5G & 1126 \\
  \rowcolor{Gray} ToMe@16, off-the-shelf &
        80.4 & 8.8G$_{\color{lightgreen}\downarrow 49.7\%}$ & 1980$_{\color{lightgreen}\uparrow 75\%}$ \\
  \rowcolor{Gray} ToMe@16, fine-tuned &
        82.8 & 8.8G$_{\color{lightgreen}\downarrow 49.7\%}$ & 1980$_{\color{lightgreen}\uparrow 75\%}$ \\
  \rowcolor{Gray} {\color{lightgray}DiffRate, fine-tuned}$\dag$ &
        83.2 & 11.5G$_{\color{lightgreen}\downarrow 34.3\%}$ & 1400$_{\color{lightgreen}\uparrow 24\%}$ \\
  \sfb/, 80 tokens$\uparrow$ &
        83.4 & 6.2G$_{\color{lightgreen}\downarrow 64.6\%}$  & 2080$_{\color{lightgreen}\uparrow 84\%}$ \\
  \sfb/, 80 tokens$\uparrow$, 320px$\uparrow$ &
        83.7 & 6.3G$_{\color{lightgreen}\downarrow 64.0\%}$  & 1898$_{\color{lightgreen}\uparrow 68\%}$ \\
  \sfb/, 100 tokens$\uparrow$, 320px$\uparrow$ &
        84.0  & 7.8G$_{\color{lightgreen}\downarrow 55.4\%}$  & 1579$_{\color{lightgreen}\uparrow 40\%}$ \\
  \hline
  \rowcolor{Gray}\augregl/ &
        85.8 & 61.6G & 388 \\
  \rowcolor{Gray} ToMe@12, off-the-shelf  &
        20.6$\ddag$ & 20.8G$_{\color{lightgreen}\downarrow 66.2\%}$ & 1006$_{\color{lightgreen}\uparrow 159\%}$ \\
  \rowcolor{Gray}ToMe@8, off-the-shelf  &
        83.5 & 31.0G$_{\color{lightgreen}\downarrow 49.7\%}$ & 700$_{\color{lightgreen}\uparrow 80\%}$ \\
  \rowcolor{Gray}ToMe@8, fine-tuned  &
        84.8 & 31.0G$_{\color{lightgreen}\downarrow 49.7\%}$ & 700$_{\color{lightgreen}\uparrow 80\%}$ \\
  \rowcolor{Gray} {\color{lightgray}DiffRate, fine-tuned}$\dag$ &
        85.7 & 42.3G$_{\color{lightgreen}\downarrow 31.3\%}$ & 452$_{\color{lightgreen}\uparrow 16\%}$ \\
  \sfl/, 80 tokens$\uparrow$ &
        85.2 & 18.6G$_{\color{lightgreen}\downarrow 69.8\%}$  & 1008$_{\color{lightgreen}\uparrow 160\%}$ \\
  \sfl/, 80 tokens$\uparrow$, 320px$\uparrow$ &
        85.5 & 18.7G$_{\color{lightgreen}\downarrow 69.6\%}$  & 928$_{\color{lightgreen}\uparrow 139\%}$ \\
  \sfl/, 100 tokens$\uparrow$, 320px$\uparrow$ &
        85.7  & 23.4G$_{\color{lightgreen}\downarrow 62.0\%}$  & 752$_{\color{lightgreen}\uparrow 94\%}$ \\
  \hline
  \rowcolor{Gray}\augreglup/ &
  86.9 & 191G & 105 \\
  \rowcolor{Gray}ToMe@40, off-the-shelf &
        8.8$\ddag$ & 56.0G$_{\color{lightgreen}\downarrow 70.7\%}$ & 315$_{\color{lightgreen}\uparrow 200\%}$ \\
  \rowcolor{Gray}ToMe@23, off-the-shelf &
        86.2 & 96.4G$_{\color{lightgreen}\downarrow 49.5\%}$ & 187$_{\color{lightgreen}\uparrow 78\%}$ \\
  \rowcolor{Gray}ToMe@23, fine-tuned &
        86.8 & 96.4G$_{\color{lightgreen}\downarrow 49.5\%}$ & 187$_{\color{lightgreen}\uparrow 78\%}$ \\
  \sfl/, 196 tokens$\uparrow$ &
        86.7 & 46.5G$_{\color{lightgreen}\downarrow 75.7\%}$ & 378$_{\color{lightgreen}\uparrow 260\%}$ \\
  \end{tabular}
  \caption{
  Comparisons on SparseFormers with token reduction methods, ToMe~\cite{tome} and DiffRate~\cite{diffrate}. $\dag$ use MAE~\cite{mae} pre-trained weights. $\ddag$ ToMe is not unfriendly to large token merging rates.
  }\label{tab:tokendropping}
\end{table}
}

\newcommand{\TableAblation}{
\begin{table}[b]
  \centering
  \footnotesize
  \renewcommand\arraystretch{1.1}
  \setlength{\tabcolsep}{2pt}
  \begin{tabular}{c|cc|cc|c}
  bootstrapping
  & \makecell[c]{with \\ cls loss} 
  & \makecell[c]{w/o reusing\\ weights} 
  & \makecell[c]{KL distill\\$\tau=3$} 
  & \makecell[c]{KL distill\\$\tau=5$} 
  & \makecell[c]{supervised cls\\ from scratch} 

        \\\shline
  82.5 & 77.4 & 69.5 & 81.0 & 81.5 &  80.0
  
  \end{tabular}
  \caption{
      Ablations on bootstrapping \sfb/ with 49 tokens.
      Except for `from scratch' (supervised classification, from scratch, 300ep), we keep the bootstrapping training settings the same (\eg, trainable parameters, epochs \& augmentation).
  }\label{tab:ablation}
\end{table}
}

\newcommand{\TableSegmentor}{
\begin{table}[t]
  \centering
  \scriptsize
  \renewcommand\arraystretch{1.1}
  \setlength{\tabcolsep}{2pt}
  \begin{tabular}{l|ccclc}
  model 
        & mIoU
        & \#Params
        & FLOPs
        & train mem.
        & img/s
        \\\shline
  Swin-L~\cite{swin} + UPerNet~\cite{upernet}$\dag$ & 51.1 & 234M & 408G & 11G, 2bs & 20 \\
  ViT-Adapter-B$_{\textrm{AugReg}}$~\cite{denseadapter} + UPerNet~\cite{upernet} & 51.9 & 134M & 376G & 50G, 2bs & 19\\
  {ViT-Adapter-L$_{\textrm{AugReg}}$~\cite{denseadapter} + UPerNet~\cite{upernet}} & 53.4 & 364M & 667G & 49G, 2bs$\ddag$ & 12 \\
  \hline
  \sfb/ segmentation (ours) & 49.3 & 68M & 36G & 10G, 4bs & 88 \\
  \sfl/ segmentation (ours) & 51.5 & 216M & 77G & 13G, 4bs & 46

  \end{tabular}
  \caption{
      Segmentation performance on ADE20k val set~\cite{ade20k}. $\dag$ is reported by mmsegmentation~\cite{mmseg2020}. $\ddag$ is with gradient checkpointing. Throughput is measured with batch size 4 on A5000.
  }\label{tab:segmentor}
\end{table}
}

\newcommand{\TableCLIP}{
\begin{table*}[t]
  \centering
  \footnotesize
  \renewcommand\arraystretch{1.1}
  \begin{tabular}{l|ccccccc}
      method 
        & \makecell[c]{visual\\encoder}
      & \makecell[c]{visual\\FLOPs}
      & \makecell[c]{IN-1K\\ zero-shot@1 }
      & \makecell[c]{MS COCO\\ I$\rightarrow$T@1}
      & \makecell[c]{MS COCO\\ T$\rightarrow$I@1}
      & \makecell[c]{Flickr30k\\ I$\rightarrow$T@1}
      & \makecell[c]{Flickr30k\\ T$\rightarrow$I@1}
      \\\shline
      \rowcolor{Gray} CLIP & ViT-B/16-224 & 17.5G & 68.3 & 52.4 & 33.0 & 81.9 & 62.0\\
      CLIP & ViT-B/32-224 & 4.4G & 63.3 & 50.1 & 30.4 & 77.5 & 58.8 \\
      \color{lightgray} TinyCLIP~\cite{tinyclip}$\dag$ &\color{lightgray} ViT-63M/32-224 &\color{lightgray} 2.0G &\color{lightgray} 64.5 &\color{lightgray} 56.9 &\color{lightgray} 38.5 &\color{lightgray} 84.9 &\color{lightgray} 66.0 \\
      CLIP + SF (\textbf{ours}) & SF-B$_{\textrm{CLIP}}$ w/ 49 tokens & 3.8G & 66.0 & 47.7 & 31.5 & 76.3 & 59.6 \\
      \hline
      \rowcolor{Gray} CLIP & ViT-L/14-224 & 81.1G & 75.4 & 56.3 & 36.5 & 85.1 & 65.2\\
      CLIP + SF (\textbf{ours}) & SF-L$_{\textrm{CLIP}}$ w/ 64 tokens & 14.8G & 73.6 & 52.3 & 34.8 & 80.3 & 62.8\\
      CLIP + SF (\textbf{ours}) & SF-L$_{\textrm{CLIP}}$ w/ 128 tokens & 30.0G & 74.9 & 54.7 & 36.4 & 82.4 & 65.2\\
      \hline
      \rowcolor{Gray} CLIP & ViT-L/14-336
      & 191.0G & 76.6 & 57.9 & 37.1 & 87.4 & 67.3 \\
      CLIP + SF (\textbf{ours}) & SF-L$_{\textrm{CLIP}}$ w/ 144 tokens
      & 33.9G & 75.9 & 57.0 & 38.3 & 84.6 & 67.7\\

      \end{tabular}
  \caption{
      Zero-shot and retrieval evaluation results. We individually bootstrap SparseFormers from CLIP visual encoders in the grayed rows with $<$2M ImageNet-1K images as inputs. $\dag$ needs training on text-image pairs of LAION-400M~\cite{laion400m}. 
  }\label{tab:clip}
\end{table*}
}

%% file: section/intro.tex
\section{Introduction}
\begin{figure}[h]
    \centering
    \includegraphics[width=\linewidth]{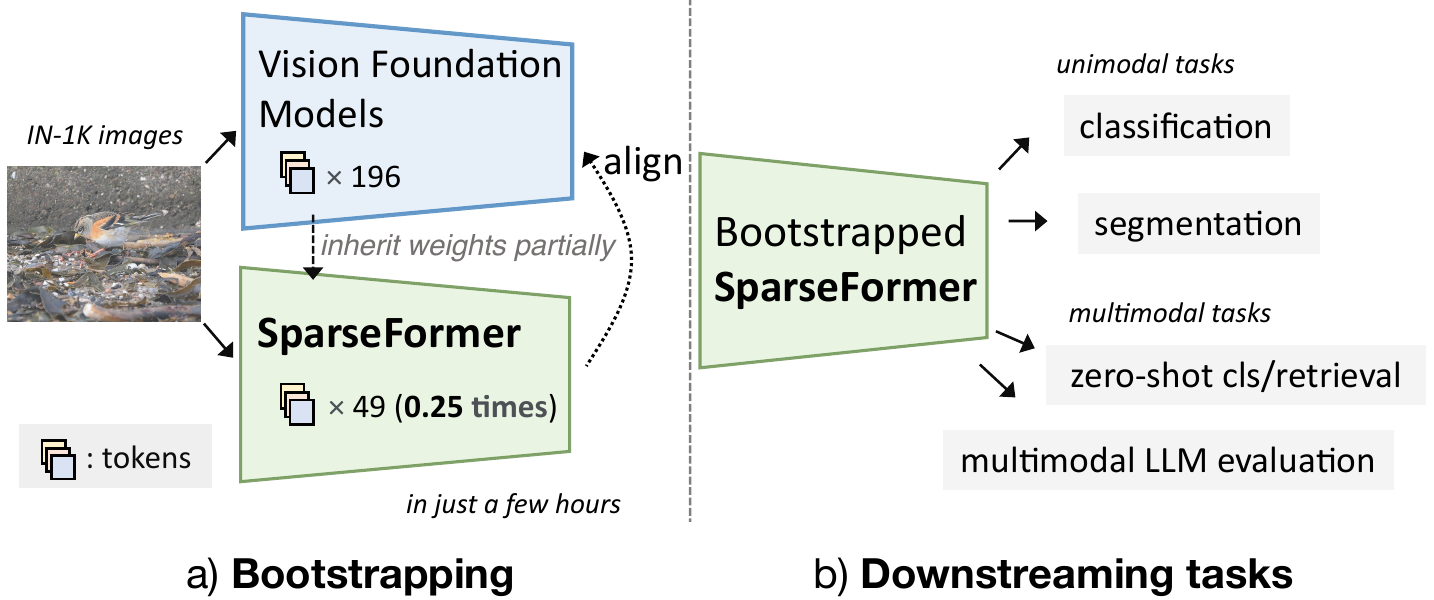}
    \caption{SparseFormer bootstrapping procedure and task evaluation.
    a) With only images as inputs, we bootstrap SparseFormers from vision foundation models by inheriting weights and aligning final representations with much fewer tokens (\eg, $0.25\times$). b) Bootstrapped SparseFormers can serve as the efficient vision encoder in either off-the-shelf or fine-tuning manner for both unimodal and multimodal tasks.} 
    \label{fig:teaser}
\end{figure}

Large-scale pre-trained vision models~\cite{zhai2022scaling,vit-22b,align,clip,augreg}, or vision foundation models, exhibit strong transferable or zero-shot capabilities after pre-training.
Most vision foundation models, e.g., ViTs~\cite{vit}, are simply based on the standard transformer encoder architecture~\cite{bert}, which is known for its high capacity and proven effectiveness in modeling massive natural language corpora~\cite{gpt3, llama}.
This large capacity is also observed in the vision domain. For instance, the large-scale ViT pre-trained on the JFT-300M dataset~\cite{jft300m} delivers notable performance on various down-streaming tasks and scaling up such models and data still consistently leads to consistent improvements~\cite{zhai2022scaling,vit-22b, clip}.
Despite their improved performance, these vision foundation transformers require significantly more computational resources and memory, both in the training and inference stage, especially when working with high-resolution images, due to the transformer architecture.
As an example, processing a single $384\times 384$ resolution image with ViT-L/16 requires handling 576 visual tokens, and attention operators between these tokens take up memory and computation quadratically with the number of tokens.

Recently, SparseFormer~\cite{sparseformer} has been proposed as an alternative vision transformer architecture with much fewer visual tokens in the latent space rather than the original image space.
Each token is associated with a region of interest (RoI) descriptor, and SparseFormer exploits the recurring focusing transformer to adjust token RoIs and sample image features sparsely according to these RoIs.
This design allows tokens to be deformable and adjustable in terms of their spatial locations.
By training with only classification labels, these latent tokens can focus on foreground objects and exclude non-informative backgrounds.
Therefore, the number of tokens in SparseFormer can be greatly reduced compared to conventional vision transformers, as well as the computational cost and memory footprint.
Though being effective in the low compute region, SparseFormer is found to be hard to scale up, and the largest SparseFormer variant in only achieves 84.8\% top-1 acc on ImageNet 1K~\cite{sparseformer}.
Moreover, despite tokens being reduced, training SparseFormers from scratch is also time-costing on the ImageNet dataset, \eg, $\sim$12 A100 GPU days for a base variant, not to mention the large-scale vision language training.
Therefore, it is interesting to explore how to efficiently train SparseFormers to serve as efficient visual transformer alternatives but with capabilities as strong as possible.

In this paper, we present a simple solution to this question: to ``bootstrap'' SparseFormers from large-scale pre-trained vision foundation models, \eg, AugRegs~\cite{augreg} and CLIPs~\cite{clip}.
By ``bootstrapping'', we mean to firstly inherit large-scale pre-trained weights from foundation models into the standard transformer encoder blocks in SparseFormers.
And then, we randomly initialize the focusing transformer in SparseFormers and train SparseFormers to align the final representation with foundation models with much fewer tokens, as depicted in Figure~\ref{fig:teaser}. 
For preserving the structure of the output space, we only tune a moderate early blocks with pre-trained weights.
Thanks to the SparseFormer efficiency and reuse of pre-trained parameters, we can quickly bootstrap scaled-up SparseFormer variants in few hours.
For instance, bootstrapping from AugReg-ViT-L/16 ($85.8\%$ on IN-1K with $388$ imgs/s) only takes just 6 hours on 8 A5000s, and the resulted SparseFormer achieves $84.5\%$ IN-1K accuracy, using only 49 visual tokens, with $1557$ imgs/s throughput.
Continuing bootstrapping with 80 tokens leads to $85.5\%$ accuracy, only $0.3\%$ lagging behind AugReg-ViT-L/16 but with $2.4\times$ throughput.
Bootstrapped SparseFormers can also serve as backbones for semantic segmentation, reaching $51+$ mIoU on ADE20k~\cite{ade20k} via 256 tokens for a 512$\times$512 input.

Moreover, since the bootstrapping procedure only needs images as inputs without any labels, we can also bootstrap SparseFormers from CLIP models to output the visual embedding in the language-aligned space just on ImageNet-1K.
Without seeing any caption, SparseFormer bootstrapping from CLIP ViT-L/14-336 demonstrates $75.9\%$ ImageNet-1K zero-shot accuracy with only 0.25$\times$ tokens, as well as the $57.0\%$ I$\to$T@1 retrieval score on the out-of-domain MS COCO~\cite{mscoco}.
In addition to that, we can also incorporate our bootstrapped SparseFormers into multimodal large language models (MLLMs) seamlessly without further fine-tuning and obtain promising results on the multimodal question answering ScienceQA dataset~\cite{scienceqa}.

%% file: section/related.tex
\section{Related work}
\subsection{Vision Foundation Models}

The term ``foundation model'' was first introduced in~\citet{bommasani2021opportunities} to refer to highly capable language transformers that were once pre-trained on massive data samples and can be easily adapted to various downstream tasks.
Exemplary of such models includes BERT~\citep{bert} and GPTs~\citep{gpt1,gpt2,gpt3}.
The concept ``foundation model'' also applies to computer vision area as well, dating back to the previous cornerstone vision backbones~\cite{vgg,resnet} pre-trained on the ImageNet.
Recent work~\citep{vit,swin,bit,augreg, sam} on computer vison focuses on pre-training models on large-scale datasets, such as ImageNet-21K~\citep{imagenet} and JFT-300M~\citep{jft300m} with the classification supervision.
The reliance on human annotations of these datasets somewhat imposes a constraint on pre-training such foundation models and several work explores contrastive learning~\citep{simclr,byol,moco}, masked modeling~\citep{mae,videomae,weakmae,eva} or other self-supervised learning~\citep{dino,dinov2} to alleviate this issue.
Despite the hunger for labels, training time and hardware demands make training a vision foundation model a difficult and expensive thing.

Data in various domains, such as web images~\cite{laion400m}, is naturally multimodal in language and vision, which invokes interests in pre-training vision models with languages~\citep{albef, blip, regionclip, unicl, icar, tcl,eva-clip}.
For example, CLIPs~\citep{clip} as the most canonical multimodal models align the output embedding spaces of the language transformer and the vision transformer, leading to remarkable zero-shot and transferring capabilities on various vision-language understanding tasks. 
The success of CLIP has also benefited many downstream tasks, including text-to-image generation~\citep{dalle2,latent_diffu,imagen} and open-vocabulary detection~\citep{owl-vit,glip,detic,f-vlm}.
Furthermore, given the growing influence and availability of large language models (LLMs)~\citep{gpt4, llama}, recent studies~\citep{blip, blip2, minigpt, llava, instructblip} aim to build multimodal models with image inputs by using CLIP pre-trained vision transformers.
But still, training these CLIP vision encoders from scratch can be extremely expensive, and such dense transformers can be computationally demanding during inference, particularly for inputs with high resolution.

\subsection{Efficient Vision Foundation Models}
It is always an appealing topic to build efficient vision transformers, \eg, with efficient attention mechanisms~\citep{wang2020linformer,kitaev2020reformer,choromanski2020rethinking}, or using compact transformer architectures~\citep{swin,heo2021rethinking,pvt,mvit}.
Beside these, a research line aims to investigate the token redundancy in vision and thus alleviate it~\citep{dynamicvit,rao2021dynamicvit,evit,avit,ats,tome,diffrate} to expedite the inference phase.
Knowledge distillation on~\citep{distilling,improved,training,beyer2022knowledge,tinyvit} have also been widely applied to architecting small transformers with knowledge transferred from large vision transformers.
Apart from unimodal ones, there has been an increasing emphasis on efficient multimodal transformers~\citep{clipping,upop,tinyclip}.

The recently proposed SparseFormer~\cite{sparseformer} as a vision transformer variant exploits the token adjusting mechanism with much fewer tokens to expedite the visual understanding task.
SparseFormer can be also seen as a vision transformer with the token reduction right from the start, but also requires lengthy training from scratch.
In this paper, we mainly discuss how to bootstrap SparseFormers from vision foundation models with limited training time and data samples, and use the bootstrapped SparseFormers as the efficient vision foundation models with most preserved performance in both unimodal and multimodal settings.

%% file: section/method.tex
\section{Method}
In this section, we will first briefly revisit the SparseFormer architecture~\cite{sparseformer}.
Next, we will describe in detail bootstrapping SparseFormers from vision foundation models with the limited training time, hardware budgets, and data samples.
\begin{figure*}
    \centering
    \includegraphics[width=0.7\linewidth]{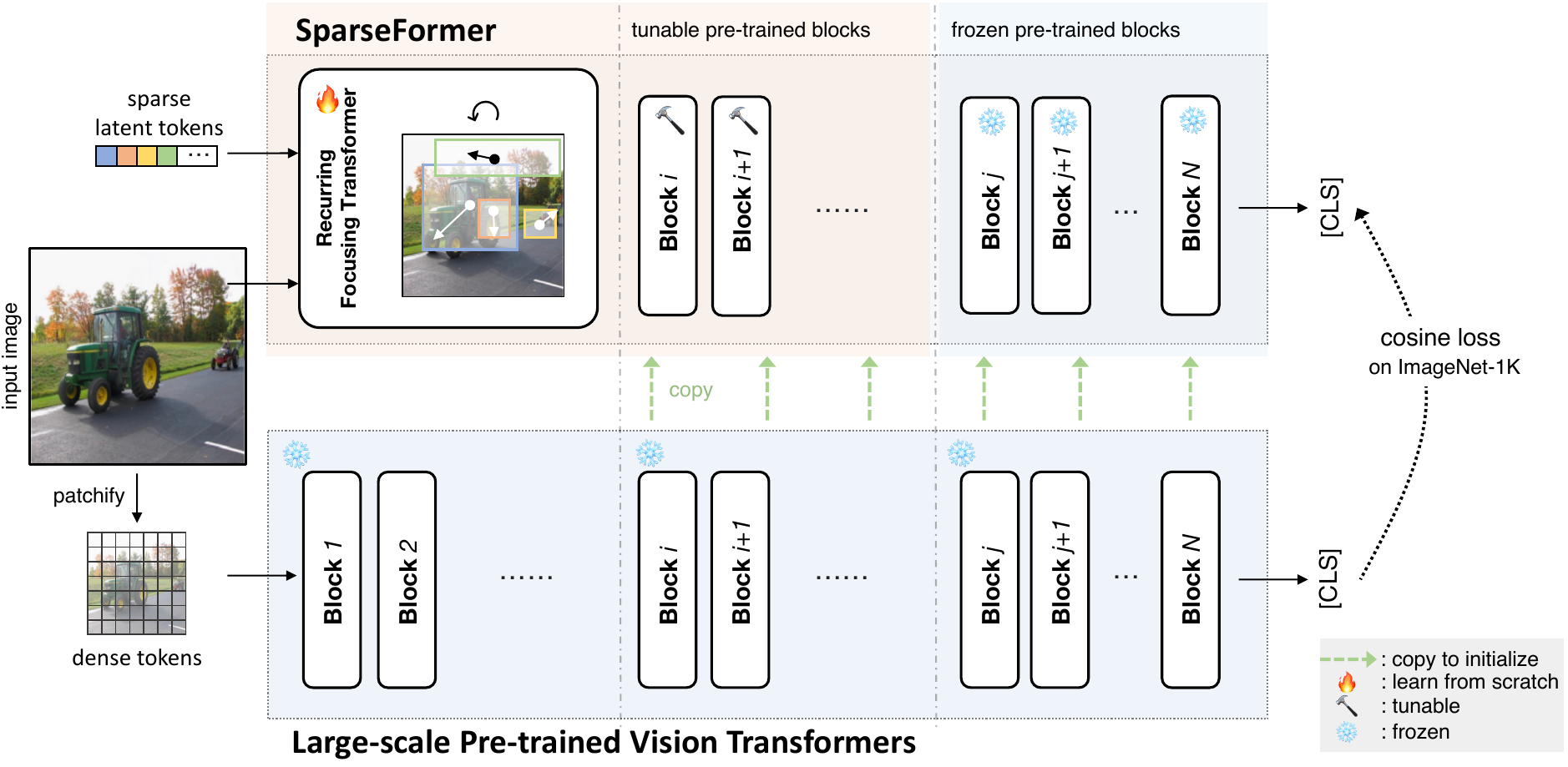}
    \caption{The detailed bootstrapping procedure.
    We typically set the number of sparse latent tokens in SparseFormers to $1/4$ those in vision transformers.
    The starting index of tunable blocks $i$ is $N/3$ and the frozen is $2N/3$ for all bootstrapping settings.
    Note that \texttt{[CLS]} represents the extra token in vision transformers besides visual tokens and there is no classification supervision on \texttt{[CLS]} in bootstrapping.}
    \label{fig:enter-label}
\end{figure*}
\subsection{Prerequisites: the SparseFormer Architecture}
The SparseFormer architecture, as a vision transformer variant, aims to represent an image by a highly reduced number of tokens and their corresponding adjustable RoIs.
SparseFormer consists up of two components: the focusing transformer and the cortex transformer.
The focusing transformer, which is designed to be with minimal parameters and computational costs, iteratively adjusts token RoIs to focus on foregrounds.
The cortex transformer is exactly a plain vision transformer encoder, similar to~\cite{vit}, which is with the majority of parameters and computation used by SparseFormer.
Specifically, given the latent token embedding set $\mathbf{T}=[\mathbf{t}_1, \mathbf{t}_2, \cdots, \mathbf{t}_N]$ and the token RoI set $\mathbf{B}=[\mathbf{b}_1, \mathbf{b}_2, \cdots, \mathbf{b}_N]$, where $N$ is the number of latent tokens, the focusing transformer operates as follows:
\begin{align}
    \mathbf{T}^{i+1}, \mathbf{B}^{i+1} = \mathrm{FocusingTransformer}(\mathbf{T}^{i}, \mathbf{B}^{i}),
\end{align}
where $\mathbf{T}^1$ and $\mathbf{B}^1$ are initial token embedding set and initial RoI set, and they are learnable parameters of models, independent of input images.
In the typical setting, the focusing transformer is repeated in 4 iterations.
In each iteration, the focusing transformer extracts image features according the current token RoI $\mathbf{b}_{(\cdot)}=[x, y, w, h]$ into token embedding $\mathbf{t}_{(\cdot)}$, perform self-attention and feed forward network on embeddings, and adjust token RoIs. The RoI adjusting mechanism uses the normalized delta form:
\begin{align}
    x' = x + \Delta_x w&, y' = y + \Delta_y h, \\
    w' = w \cdot \exp(\Delta_w)&, h' = h \cdot \exp(\Delta_h),
\end{align}
where the tuple $(\Delta_x, \Delta_y, \Delta_w, \Delta_h)$ is produced by the token embedding in the focusing transformer.
All operations are applied individually to each token, except for self-attention which involves token interaction.

After the focusing transformer, these token RoIs tend to cluster around foreground objects~\cite{sparseformer}.
Then, the so-called cortex transformer, whose architecture is exactly a standard transformer encoder, operates on the token embedding as:
\begin{align}
    \mathbf{T}^{\mathrm{final}} = \mathrm{CortexTransformer}(\mathbf{T}'),
\end{align}
where we can see that the cortex transformer just processes token embeddings and does not rely on the token RoIs anymore.
With the RoI adjusting mechanism, SparseFormer utilizes a highly reduced number of tokens (e.g., $0.25\times$) for visual understanding by only attending to foreground objects and excluding backgrounds.

\paragraph{Feasibility to bootstrap SparseFormers from ViTs.}
Since the cortex transformer strictly follows the standard transformer encoder architecture, it is feasible to inherit pre-trained vision transformer weights.
Once the weights of the cortex transformer are inherited, we only need to train the focusing transformer from scratch to align its representation into the cortex transformer to be compatible with inherited cortex transformer weights.
Thus, when the input representation well aligned, the output representation of SparseFormer will also align with the original pre-trained vision transformers, which is especially beneficial for language-aligned vision representations, such as CLIPs~\cite{clip}.
We term this procedure as ``bootstrapping'' SparseFormers from large-scale pre-trained vision foundation models.

\subsection{Bootstrapping from Vision Foundation Models}
As discussed above, we need to align the output representation of the focusing transformer with the input representation of inherited vision transformers.
However, this straightforward idea is infeasible since the shape of these two representations typically does not match as the number of SparseFormer tokens is much lower than ones in most vision transformers.

Therefore, we resort to directly aligning the final representation of SparseFormer to one from pre-trained vision transformers.
Specifically, we propose a simple method to bootstrap SparseFormers from vision foundation models, which are mostly based on transformers.
First, we design SparseFormer variants to match the dimension and the number of their cortex transformer blocks with these pre-trained vision transformers.
We inherit these large-scale pre-trained weights of these transformers and load them into cortex transformer blocks in SparseFormers.
Then, we initialize the focusing transformer from random and concatenate these two to build a complete SparseFormer.

Once a SparseFormer is constructed, we align the final representation from SparseFormer with that of the pre-trained vision transformer given the same image input, as depicted in Figure~\ref{fig:enter-label}.
We simply use the cosine loss to align these two final token representations:
\newcommand{\qst}{\mathbf{t}_{\mathrm{sparseformer}}}
\newcommand{\qtt}{\mathbf{t}_{\mathrm{pretrained}}}
\begin{align}
    \ell = 1 - \frac{\qst^T\qtt}{\|\qst\|\|\qtt\|},
\end{align}
where $\qst$ and $\qtt$ are the final token embeddings in the last embedding space of models, which are \texttt{[CLS]} tokens before the classifier layer in classification models or after the projection layer in CLIP models.
We can bootstrap SparseFormers from large-scale pre-trained vision foundation models without labels, since the alignment target from pre-trained models, $\qtt$, already has rich semantics.
All we need to do is to seek for a set of unlabeled supporting images.
We find that the medium-sized ImageNet-1K training set~\cite{imagenet}, which is most publicly available to its scale, with minimal augmentation is a sufficient supporting image set for bootstrapping competitive SparseFormers from pre-trained models.

\paragraph{Difference with distillation methods.}
It is important to note that the cosine loss between two token embeddings is not dependent on pre-defined label sets, and we do not exploit any labels or captions during the bootstrapping process.
This differs a lot from existing distillation methods, which typically make use of specified label sets~\cite{distilling, training} for classification models or vision-language pairs for multimodal models~\cite{tinyclip}.
Also, distillation methods do not reuse teacher model weights since the student architecture usually differs from the teacher, and thus require a lengthy schedule to transfer knowledge effectively.
In comparison, our method can bootstrap a 200M SparseFormer variant in a few hours ($20$ epochs) with only 1.2M images from ImageNet as inputs.

\paragraph{Truncate the leading, tune the middle, and freeze the ending blocks.}
The leading blocks of pre-trained vision transformers are usually specialized for low-level visual modeling~\cite{vitseelike}, which overlaps with the role of the focusing transformer in SparseFormers.
Thus, we choose to discard a moderate number, $1/3$, of leading transformer blocks when constructing our SparseFormer variants, which also help reduce the compute need.
Since the bootstrapping goal is to well align the final representation, we set a few middle transformer blocks tunable starting from inherited weights to adapt the output of the focusing transformer.
We leave frozen the rest transformer blocks in the ending to preserve the structure of the output space by pre-trained weights as much as possible.

%% file: section/exps.tex
\section{Experiments}
We start experiments on bootstrapping SparseFormers from large-scale pre-trained unimodal models and then discuss vision-language pre-trained ones.
\subsection{Bootstrapping from Unimodal Models}~\label{sec:unimodal}
We choose the well-established AugReg vision transformers~\cite{augreg}, which strictly follow the original ViT architecture, as the unimodal classification models to bootstrap from.
The AugReg models are pre-trained by the supervised classification on the ImageNet-21K dataset with well-curated data augmentation and regularization techniques.

\paragraph{Model configurations and experimental settings.}~\label{subsec:settings}
We design two SparseFormer variants (\sfb/ and \sfl/) according to the base and large variant (\augregb/ and \augregl/) in Table~\ref{tab:configuration}. 
We set the token dimension in the focusing transformers half those of the cortex transformers except that the last focusing transformer block performs sampling operations on the full dimension, in accord with \cite{sparseformer}.
Unlike the leading one, the last focusing transformer block does not involve FFNs and self-attention layers.
We append a learnable \texttt{[CLS]} to latent tokens after the focusing stage in SparseFormer and use its final representation for bootstrapping and classification.
To be more consistent with AugReg models (CLIPs below as well) which add positional encoding to visual tokens, we inject positional encoding based on token RoIs in the sinusoidal form in each focusing iteration.
To further reduce compute and make SparseFormer even more sparse, we reduce the channel of the early convolution from $96$ to $64$, and shrink the number of sampling points for a token in the sparse sampling procedure from $36$ to $16$.

\TableConfiguration
We choose the publicly-available medium-sized ImageNet-1K training set as our bootstrapping image set.
The data augmentation is set to be minimal: random horizontal flipping with the probability $0.5$ and random resized cropping with from the scale $(0.5, 1.0)$ and the aspect ratio $(3/4, 4/3)$.
We use a budget of $20$ epochs as our bootstrapping schedule, where learning rate begins at $2\times 10^{-4}$ and follows a half-cosine decay schedule after the first warming-up epoch.
In the warming-up epoch, the inherited pre-trained weights keep frozen for training stability.
In bootstrapping, only the focusing transformer and tunable blocks of cortex transformer are learnable.
All the rest parameters inherited from AugReg models, including the final layer normalization and classification layer, remain frozen.

\TableUnimodalExp
\paragraph{Bootstrapped SparseFormers with 49 tokens.}
We show results of bootstrapped SparseFormer from AugReg classification models in Table~\ref{tab:unimodal}.
The default token number is 49 for SparseFormer models.
From the results, we can see that SparseFormer models with 49 tokens have achieved promising results, \eg, 84.9\% top-1 accuracy for \sfl/ bootstrapping from \augreglup/.
Note that the number of tokens, \eg 49, is fixed from the beginning to the end in SparseFormer models, which differs from recent other methods progressively dropping tokens through blocks~\cite{tome,diffrate}.
To our best knowledge, we are the first to obtain $\sim$85\% top-1 accuracy on ImageNet-1K with only 49 tokens in all blocks for vision transformers.

\paragraph{Continue to bootstrapping with more tokens.}
Although SparseFormers with 49 tokens yield promising results, they are still lagging behind those models to bootstrap from with a non-negligible margin.
To minimize the gap, we further continue to bootstrapping SF models with the increased number of tokens from default SF models.
Note that these increased numbers are still much lower than original vision transformers, especially for high resolution one.
The continued bootstrap lasts for 5 epochs with LR starting from $5\times10^{-5}$, also following a half-cosine decay scheme.
We denote these models with an extra notation $\uparrow$.
From the table, the gaps between SparseFormers and AugRegs become closer when original models become larger, especially for \sfl/ with 196 tokens, which is on par with \augreglup/ but at about 3$\times$ real-time throughput.
When combined with the higher input resolution, SparseFormer models still yield improvements on the same number of tokens at a marginal additional burden.
Note that as a sparse architecture, the input resolution has little impact on computational costs of the most components in SparseFormers.

\TableAblation
\paragraph{Ablation studies on bootstrapping.}
Our bootstrapping procedure is simple as it just aligns final representations of SparseFormers and large-scale pre-trained models given the same image input.
Labels are not needed through the bootstrapping procedure except for the evaluation on benchmarks.
Here we ablate the bootstrapping design in Table~\ref{tab:ablation}.
Interestingly, aligning the final representations along with the classification loss (`with cls loss' in the table) actually impairs the performance.
We suspect that the supervised classification requires strong augmentation and regularization to be stable with the alignment.
We also replace our token representation alignment with KL distillation loss with different temperatures on the ImageNet labels.
This simple distillation yields promising results but still with $>1\%$ lagging behind our bootstrapping.

We also include the bootstrapping result without the reusage of pre-trained weights (but the final linear classifier remained) and train the entire transformer to align the final representation for 20 epochs.
As expected, it yields the inferior result, and more training epochs may be required to well align from scratch.

\paragraph{Comparison with token reduction methods.}
\TableComparisonWithTokenDropping
SparseFormers can be considered as a method to reduce the number of visual tokens in vision transformers after pre-training, similar to token pruning or merging approaches~\cite{evit, tome, ats, diffrate}.
Here, we investigate the effectiveness of SparseFormers from the token reduction perspective and compare them to state-of-the-art reduction methods, ToMe~\cite{tome} and DiffRate~\cite{diffrate}, in Table~\ref{tab:tokendropping}.
In addition to using off-the-shelf AugReg models with ToMe, we also fine-tune AugRegs with ToMe using the fine-tuning recipe described in~\cite{mae} to ensure a fair comparison with our methods, as additional fine-tuning is required.
We sweep to find best fine-tuning LRs for these ToMe AugRegs respectively.
DiffRate on MAE models are also included, where the \mael/ baseline reports a slightly better result than \augregl/.
As shown in the table, SparseFormer models achieve the best trade-off between the accuracy and the actual throughput, especially for the large models where many tokens may be redundant.
In fact, tokens in bootstrapped SparseFormers can be further applied with these token reduction methods but this is beyond this paper scope.

\paragraph{Dense prediction task.}
\TableSegmentor
\TableCLIP
We here further investigate the ability of bootstrapped SparseFormers as pre-trained backbones to perform dense per-pixel prediction task, \eg, semantic segmentation.
As latent tokens of SparseFormer are not structured in a grid-like map like conventional vision transformers~\cite{vit,swin,pvt}, it requires some workarounds to obtain dense feature maps from these latent tokens.
We generally follow the idea of the original SparseFormer on semantic segmentation to project latent tokens back into the original dense pixel space according to their RoIs~\cite{sparseformer}.
Here, we find that it is more effective to straightforward project prediction logits of tokens, instead of embeddings, back into the dense pixel space using simple single-head attention aggregation:
\begin{align}
    \mathbf{P}_{\textrm{token}} &= \textrm{classifier}(\mathbf{T}) \in \mathbb{R}^{N\times L}, \\
    \mathbf{P}_{\textrm{dense}} &= \textrm{softmax}(\mathbf{Q}_{\textrm{dense}}\mathbf{K}_{\textrm{token}}^T/\sqrt{d}+\mathbf{B})\mathbf{P}_{\textrm{token}},
\end{align}
where $\textrm{classifier}(\cdot)$ just classify the latent token into prediction logits, $N$ is the number of latent tokens, and $L$ is the number of classes.
In the current implementation, we first use a linear layer to reduce the latent token dimension to $d=256$ and introduce two small transformer encoder blocks ($d=256$) to process latent tokens, and we use a linear layer as our latent classifier.
$\mathbf{Q}_{\textrm{dense}}\in\mathbb{R}^{HW/4^2\times d}$ is the flattened feature map transformed by one $3\times3$ convolution from the early convolution feature in SparseFormer with shape $H/4\times W/4$.
We bias the attention score by
$\mathbf{B}=\mathbf{B}_{\textrm{geometric}}+\mathbf{B}_{\textrm{predictive}}$, where the geometric bias $\mathbf{B}_{\textrm{geometric}}$ is in the Gaussian-like form w.r.t. token RoIs as described in \cite{sparseformer}.
The predictive bias is produced by a linear layer whose input is the latent token embedding, $\mathbf{B}_{\textrm{predictive}}=\mathrm{linear}(\mathbf{T})\in\mathbb{R}^{1\times N}$, indicating the overall importance of token logits into the dense map.
In simple words, we map the latent token predictions into the dense grid with the consideration of their RoIs and semantic importance, similar to \cite{maskformer}.
The classification supervision is imposed on the mapped dense predictions.

We generally follow the training recipe of \cite{denseadapter}, including learning rate and drop path~\cite{droppath} settings, and finetune our bootstrapped SparseFormers (default ones w/ 49 tokens) on the ADE20k dataset~\cite{ade20k} with the increased token number of 256.
As our segementation models based on SparseFormer are memory friendly, the batch size is increased to 4 and the total training iteration is reduced to 80K.
The results are reported in Table~\ref{tab:segmentor}.
We can see that the performance of SparseFormer on the segmentation task is promising, with much fewer FLOPs and much higher throughput.
This efficiency is credited to fewer computation over limited latent tokens in SparseFormers compared to conventional dense vision transformers (\eg, 256 versus 1024 for ViT-Adapter~\cite{denseadapter} for 512$\times$512 inputs).
\subsection{Bootstrapping from Multimodal Models}~\label{sec:clip}
Till now, we have bootstrapped SparseFormers from unimodal classification models and apply them to the downstreaming dense prediction task, showing their efficiency and effectiveness.
Since our bootstrap procedure directly aligns the final representations to the pre-trained transformers, it can be easily adopted for the fundamental vision-language CLIP~\cite{clip} models as well.

\paragraph{Experimental setups.}
The bootstrapping procedure for CLIP models exactly follows the same recipe in unimodal classification AugRegs, as described in Sec~\ref{sec:unimodal}, including the training budget, the data augmentation, and the learning rate setting.
We use the official OpenAI CLIP pre-trained vision transformers to bootstrap from, \ie, ViT-B/16-224, ViT-L/14-224, and ViT-L/14-336.
We only bootstrap SparseFormers from the CLIP visual encoders and leave the text encoders untouched.
It is worth noting that we still only use the ImageNet-1K training set as the supporting image dataset, and we do not leverage any text-image pairs to align the SparseFormer image embeddings with CLIP models.
This differs from TinyCLIP~\cite{tinyclip}, which also inherits pre-trained weights but requires at least 15M text-image pairs to distill smaller models.

\paragraph{Results.}
We use the frozen CLIP text encoders in combination with our bootstrapped SparseFormers as visual encoders to perform the zero-shot classification task on ImageNet-1K validation set, as well as the retrieval task on MS COCO val set~\cite{mscoco} and the Flickr30k benchmark~\cite{flick30k}.
Results are presented in Table~\ref{tab:clip}.
With only images as samples for aligning in the bootstrapped procedure, SparseFormers as the visual encoders achieve decent zero-shot accuracies with greatly reduced compute, especially for large and high resolution models.
For instance, our SparseFormer-L with 144 tokens is just 0.7 point behind the ViT-L/14-336 with 576 tokens, and surpass the ViT-L/14-224 by 0.5 margin with less than 0.5$\times$ computational cost. 
Interestingly, the zero-shot retrieval performance on MS COCO and Flickr30k, which have more complicated data distributions compared to the ImageNet samples for aligning, is also quite effective.

\subsection{Multimodal Large Language Models}
\TableScienceQA
\begin{figure*}[h!]
    \centering
    \includegraphics[width=\linewidth]{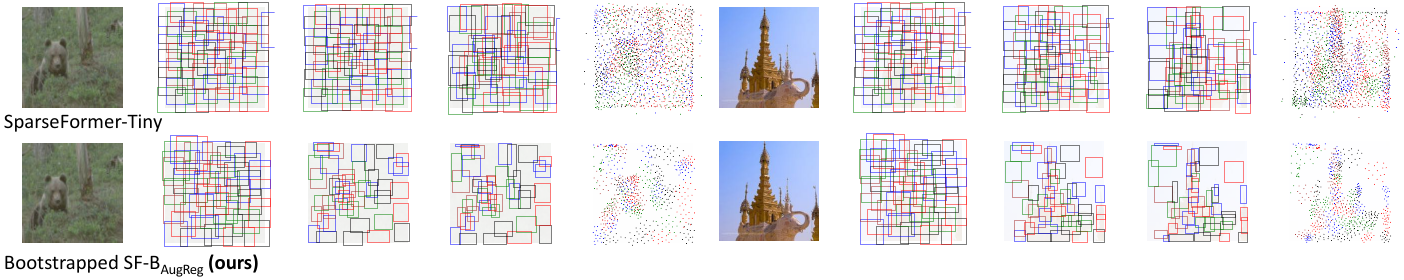}
    \caption{
        Visualization on the original SparseFormer-Tiny~\cite{sparseformer} and our bootstrapped \augregb/. For each image, there are an input image, token RoIs in the \{first, third, last\} stage, and sampling points in the last stage in the focusing transformer from left to right.
    }
    \label{fig:vis}
\end{figure*}

Building large language models with integrated vision capabilities is an emerging and appealing topic~\citep{llava,blip,blip2} within the computer vision community.
Since we have bootstrapped SparseFormers from CLIP models, a natural question arises: can we seamlessly incorporate SparseFormer as the vision encoder in the multimodal large language model without re-finetuning LLMs.
The answer is yes.
We experiment with LLaVa~\cite{llava}, a recent large language model that is fine-tuned for instructions and can accept images and texts as the prompt.
The vision encoder in LLaVA is typically the CLIP ViT-L and it directly outputs all visual tokens (typical number is 256 for a 224$^2$ image) as input tokens for autoregressive language modeling, meaning that there are already 256 ``word tokens'' for image modeling before user input prompt.
In comparison, SparseFormers reduce this number of tokens to typically $1/4\times$, which is more friendly for LLM text modeling.

To demonstrate quantitative multimodal LLM results, we benchmark the fine-tuned LLaVa variant, \texttt{llava-lcs558k-scienceqa-vicuna-13b-v1.3}, for the instruction following on the ScienceQA multi-choice question answering task~\citep{scienceqa}.
We find that the data distribution is unique on the ScienceQA dataset and directly using bootstrapped SparseFormers in Sec~\ref{sec:clip} leads to inferior results.
Therefore, we continue bootstrapping SparseFormers from CLIP ViT-L/14-224 using the ScienceQA training images for $\sim$1000 iterations to adapt to this image domain for about 20 minutes.
Then we replace the CLIP vision encoder in LLaVa with SparseFormers without any further fine-tuning LLM.
Results are shown in Table~\ref{tab:scienceqa}.
The bootstrapped SparseFormer as the vision encoder preserves most language-vision abilities ($84.5$ vs $88.6$ in the multimodal only entry) with only $1/4\times$ visual tokens as inputs to the following autoregressive LLM.
The performance gap between SparseFormers and CLIP even narrows further when scaling to 128 tokens, which is still half the size of CLIP tokens.

\subsection{Visualization}
We show the visualization of our bootstrapped SparseFormer and compare it to the original SparseFormer~\cite{sparseformer} in Figure~\ref{fig:vis}.
While the bootstrapped \augregb/ and the original SparseFormer-Tiny use exactly the same number of visual tokens (\ie, 49), our bootstrapped model demonstrates the better capability to focus on foregrounds and exclude non-informative regions with less sampling points (36$\to$16 per token, as described in Section~\ref{subsec:settings}).
As a result, the bootstrapped SparseFormer from large-scale pre-trained models can be {\em even sparser} in both focal ability and sampling efficiency compared with the original one.

%% file: section/conclusion.tex
\section{Conclusion}
In this paper, we have proposed a simple but effective method to bootstrap SparseFormers from large-scale pre-trained vision foundation transformers by inheriting most pre-trained weights and explicitly aligning the final representations.
With the short training time and limited data samples, the SparseFormers bootstrapped from unimodal classification models can maintain the pre-trained capabilities as much as possible with much fewer tokens and higher real-time throughput.
In particular, we can bootstrap a SparseFormer with only 49 tokens which obtains 84.9 top-1 accuracy on ImageNet 1K.
Also, SparseFormers can serve as backbone networks in the dense per-pixel semantic segmentation task with decent results but at 2$\times$ throughput.
In addition to classification models, SparseFormers can be bootstrapped from multimodal pre-trained transformer CLIPs without seeing any text caption in the bootstrapping procedure and the bootstrapped SparseFormers demonstrate strong multimodal performance with down to 0.2$\times$ FLOPs.
\paragraph{Limitation}
The limitation of our proposed bootstrapping procedure is that it presumes the underlying transformer architecture of vision foundation models and thus can only applied to transformer-based foundation models.
This is mostly the case, but there are some exceptions, \eg,~\cite{convnext}.
Also, it necessitates the availability of weights of vision foundation models, while that may not be possible for some large-scaled proprietary ones.

\section*{Acknowledgement}
This project is supported by the National Research Foundation, Singapore under its NRFF Award NRF-NRFF13-2021-0008.